\documentclass[11pt]{article}

\usepackage[preprint]{acl}

\usepackage{times}
\usepackage{latexsym}
\usepackage[table,dvipsnames]{xcolor}
\usepackage{colortbl}
\usepackage{enumitem}
\usepackage[most]{tcolorbox}

\definecolor{lightblue}{RGB}{220,235,250}
\usepackage[T1]{fontenc}
\usepackage[utf8]{inputenc}

\usepackage{microtype}

\usepackage{inconsolata}

\usepackage{graphicx}
\usepackage{amsmath}
\usepackage{amssymb}
\usepackage{tcolorbox}
\usepackage{longtable}
\usepackage{booktabs}
\usepackage{xcolor}
\usepackage{algorithm}
\usepackage{algorithmic}

\tcbuselibrary{listings,breakable}
\usepackage{bbding}
\usepackage{pifont}

\title{TR-ICRL: Test-Time Rethinking for In-Context Reinforcement Learning}

\author{
 \textbf{Wenxuan Jiang\textsuperscript{1,2}\thanks{~~Equal contribution.}\thanks{~~Work done during an internship at Meituan.}},
 \textbf{Yuxin Zuo\textsuperscript{3}\footnotemark[1]},
 \textbf{Zijian Zhang\textsuperscript{2}},
 \textbf{Xuecheng Wu\textsuperscript{4}},
 \textbf{Zining Fan\textsuperscript{5}},
 \\
 \textbf{Wenxuan Liu\textsuperscript{3}},
 \textbf{Li Chen\textsuperscript{6}},
 \textbf{Xiaoyu Li\textsuperscript{2}},
 \textbf{Xuezhi Cao\textsuperscript{2}},
 \textbf{Xiaolong Jin\textsuperscript{3}\thanks{~~Corresponding author.}},
 \textbf{Ninghao Liu\textsuperscript{1}\footnotemark[3]},
\\
 \textsuperscript{1}The Hong Kong Polytechnic University
 \textsuperscript{2}Meituan-M17 \\
 \textsuperscript{3}Institute of Computing Technology, Chinese Academy of Sciences
 \\
 \textsuperscript{4}Xi'an Jiaotong University
 \textsuperscript{5}East China Normal University
 \textsuperscript{6}Northeastern University
\\ \vspace{0.2cm}
 \texttt{pangxuan022@gmail.com},
 \texttt{\{zuoyuxin24s, jinxiaolong\}@ict.ac.cn},
 \texttt{ninghliu@polyu.edu.hk}
}

\begin{document}
\maketitle

\begin{abstract}
In-Context Reinforcement Learning (ICRL) enables Large Language Models (LLMs) to learn online from external rewards directly within the context window. However, a central challenge in ICRL is reward estimation, as models typically lack access to ground-truths during inference. 
To address this limitation, we propose \textbf{Test-Time Rethinking for In-Context Reinforcement Learning (TR-ICRL)}, a novel ICRL framework designed for both reasoning and knowledge-intensive tasks.
TR-ICRL operates by first retrieving the most relevant instances from an unlabeled evaluation set for a given query.
During each ICRL iteration, LLM generates a set of candidate answers for every retrieved instance. Next, a pseudo-label is derived from this set through majority voting.
This label then serves as a proxy to give reward messages and generate formative feedbacks, guiding LLM through iterative refinement.
In the end, this synthesized contextual information is integrated with the original query to form a comprehensive prompt, with the answer determining through a final round of majority voting.
TR-ICRL is evaluated on mainstream reasoning and knowledge-intensive tasks, where it demonstrates significant performance gains.
Remarkably, TR-ICRL improves Qwen2.5-7B by \textbf{21.23\%} on average on MedQA and even \textbf{137.59\%} on AIME2024.
Extensive ablation studies and analyses further validate the effectiveness and robustness of our approach.
Our code is available at \url{https://github.com/pangpang-xuan/TR_ICRL}.

\end{abstract}

\begin{figure*}[t!]
\centering
\includegraphics[width=0.80\textwidth]{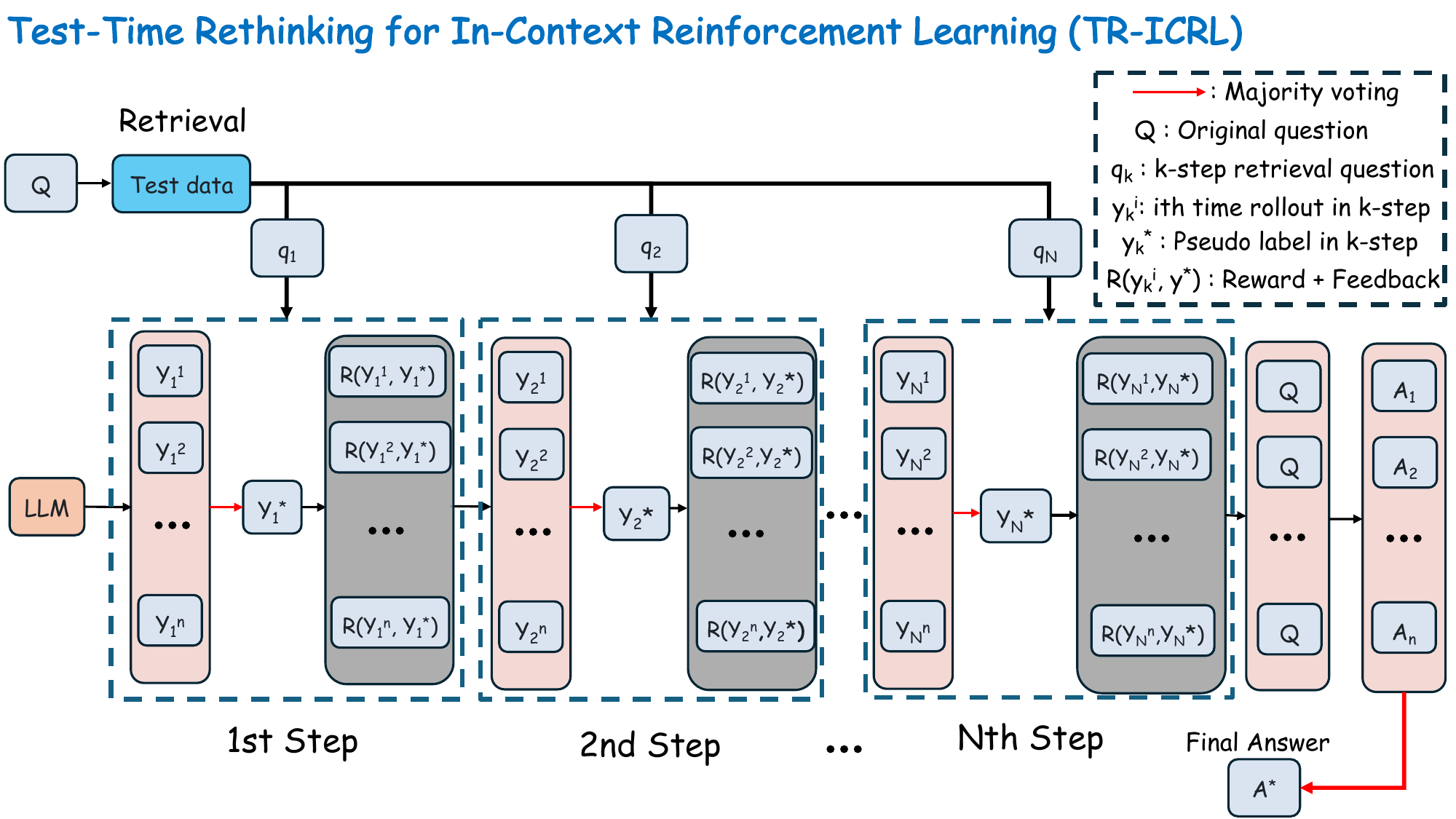}
\caption{\textbf{TR-ICRL} combines both Test-Time Scaling (TTS) and In-Context Reinforcement Learning (ICRL).}
\label{fig:framework}
\end{figure*}

\section{Introduction}
Large Language Models (LLMs)~\cite{yan2025muse, guo2025deepseek, jaech2024openai} have demonstrated remarkable advancements across a wide range of domains. Recently, OpenAI's o1~\cite{jaech2024openai} and Deepseek-R1~\cite{guo2025deepseek} have shown that Test-Time Scaling (TTS)~\cite{zhang2025and, balachandran2025inference} can significantly enhance the reasoning capabilities of LLMs by leveraging additional computational resources during inference.

Following the breakthrough of in-context supervised learning \cite{brown2020language}, there is growing interest in formalizing In-Context Reinforcement Learning (ICRL) \cite{moeini2025survey, monea2024llms, ye2026context}.
While early theoretical instantiations were predominantly confined to simulated environments such as multi-armed bandits~\cite{krishnamurthy2024can}, ICRL has rapidly evolved into a versatile paradigm for sophisticated reasoning.
Recent empirical evidence~\cite{song2025reward} demonstrates that in-context reinforcement learners exhibit a powerful emergent capability to autonomously refine their strategies across diverse, high-stakes domains. These applications now range from orchestrating complex scientific experiments and open-ended creative writing to solving olympiadlevel mathematics.
By bypassing the need for explicit gradient updates~\cite{brown2020language}, ICRL enables LLMs to adapt to the nuances of these knowledge-intensive tasks in real-time, purely through contextual interaction.

Despite these advancements, this raises a fundamental question: \textbf{how can Test-Time Scaling be optimally interleaved with in-context reinforcement learning to facilitate autonomous strategy refinement?}
A primary bottleneck in deploying ICRL within open-ended environments is its traditional dependence labeled data.
In the absence of external supervision, a fundamental challenge arises: how to derive robust reward signals during test time to facilitate self-improvement without access to ground-truth labels.

To mitigate these challenges, we introduce \textbf{Test-Time Rethinking for In-Context Reinforcement Learning (TR-ICRL)}, a novel framework that leverages Test-Time Scaling to generate self-consistent rewards, thereby bolstering the performance of In-Context Reinforcement Learning during inference, as illustrated in Figure~\ref{fig:framework}.

TR-ICRL begins by retrieving the $N$ most similar instances from the unlabeled evaluation set for a given input query.
For each retrieved instance, the model executes $K$ parallel rollouts to establish a pseudo-label via majority consensus.
Each prediction, $y_i$, is then evaluated against the pseudo-label, initiating a \textbf{rethinking} stage.
In this stage, the LLM processes a reward message to generate formative feedback, specifically identifying and rectifying latent reasoning fallacies. Once these $N$ iterations are complete, the model synthesizes the full contextual history—integrating retrieved predictions and their corrective feedback—to address the original query.
Finally, a self-consistent aggregation phase compiles all candidate answers, using majority voting to determine the final output.

In our experiments, we evaluate TR-ICRL across a diverse suite of models, including two instruction-tuned models and two Large Reasoning Models (LRMs).
We evaluate TR-ICRL on 5 reasoning benchmarks and 2 knowledge-intensive benchmarks to assess the its versatility.
Notably, integrating TR-ICRL with Qwen2.5-7B yields a substantial average improvement of \textbf{58.92\% }(increasing from \textbf{34.80 to 55.30}) on the AMC benchmark, alongside a remarkable \textbf{137.59\%} gain \textbf{(from 7.9 to 18.7)} on the more challenging AIME2024.
For knowledge-intensive tasks, Llama3.1-8B exhibits a marked improvement of \textbf{21.22\%} on MedQA and a significant \textbf{36.68\%} increase on MedXpertQA.
Our preliminary findings demonstrate that TR-ICRL is highly effective across diverse model scales and task domains.
Furthermore, we conducted ablation studies to evaluate the retrieval quality and the ranking of the selected questions.
Notably, the framework exhibits robust scalability, suggesting a high performance ceiling as model capacity increases. In conclusion, our main contributions are three-fold:
\begin{itemize}
    \item We introduce \textbf{Test-Time Rethinking for In-Context Reinforcement Learning}, a novel ICRL framework designed for both intensive reasoning and knowledge-intensive tasks.
    \item We evaluate TR-ICRL on multiple LLMs across five mathematical reasoning and two knowledge-intensive benchmarks, demonstrating consistent and significant improvements over the base models.
    \item Extensive experiemntal results demonstrate that TR-ICRL enables efficient and stable ICRL in a fully unsupervised manner, effectively eliminating the requirement for ground-truth labels during inference.
\end{itemize}

\section{Related Work}
\subsection{In-Context Reinforcement Learning}
Reinforcement Learning~\cite{zhang2025survey} has emerged as a primary catalyst for augmenting the reasoning capabilities of LLMs.
Building upon this foundation, In-Context Reinforcement Learning~\cite{moeini2025survey} has been formalized to describe models that adapt their behavior dynamically without the need for gradient updates.
A central tenet of ICRL involves conditioning a policy $\pi_\theta$ on both the current state $s_t$ and a dynamic context $C_t$, where actions are sampled according to $\pi_\theta(a_t \mid s_t, C_t)$ \cite{duan2016rl}.
While $C_t$ can be instantiated through various mechanisms, a prevalent convention defines it via the historical trajectory $\tau_t$.
The efficacy of ICRL rests on the hypothesis that the forward pass of a static neural network $\theta$ implicitly executes an RL procedure, enabling the policy to generalize to out-of-distribution Markov Decision Processes at test-time \cite{laskin2023incontext}.
This phenomenon often termed in-context improvement—posits that LLM performance scales monotonically with context length, provided $C_t$ remains semantically relevant to the underlying task.

\subsection{Test-Time Scaling}
Test-Time Scaling \cite{zhang2025and,balachandran2025inference} enhances the reasoning capabilities of LLMs during inference by leveraging additional computational resources without altering model weights. A foundational technique is CoT~\cite{wei2022chain}, which encourages models to “think step by step” \cite{lightman2023let}, significantly improving performance on complex tasks. More structured approaches include Best-of-N (BoN) sampling \cite{brown2024large}, beam search \cite{snell2024scaling}, and Monte Carlo Tree Search \cite{zhang2024rest}. These methods generate multiple candidate solutions, often applying majority voting \cite{stiennon2020learning}, PRM \cite{yuan2024free} as verifier, or LLM-as-a-judge \cite{zheng2023judging} for greater accuracy.

\section{TR-ICRL}
\begin{algorithm*}[t]
\caption{Test-Time Rethinking for In-Context Reinforcement Learning (TR-ICRL)}
\label{alg:TR-ICRL}
\small
\begin{algorithmic}[1]
    \REQUIRE Query $x_{\text{test}}$, unlabeled evaluation set $\mathcal{D}$, LLM $\pi_{\theta}$, Step $N$, Rollout number $K$, Retrieval $\mathcal{R}$.
    \ENSURE Final prediction $y^*$.
    
    \STATE $\mathcal{Q} = \{q_1, q_2, \dots, q_N\} \leftarrow \mathcal{R}(x_{\text{test}}, \mathcal{D})$
    \STATE Initialize $K$ message buffers $\mathcal{M}^{(0)} = \{m_1^{(0)}, \dots, m_K^{(0)}\}$ with system prompt.
    
    \FOR{$i=1$ \TO $N$}
        \STATE $\{a_{i}^{(1)}, \dots, a_{i}^{(K)}\} \leftarrow \pi_{\theta}(m_k^{(i-1)} \oplus q_i) \text{ for } k=1 \dots K$
        \STATE $\hat{y}_{i} \leftarrow \text{Vote}(\{ \text{Extract}(a_{i}^{(k)}) \}_{k=1}^K)$
        \STATE $\{r_{i}^{(1)}, \dots, r_{i}^{(K)}\} \leftarrow \text{Reward}
        (\text{Extract}(a_{i}^{(k)}), \hat{y}_{i})  \text{ for } k=1 \dots K$ \text{\,\,\,\,\,\,\,\,// Reward calculations.}
        \STATE $\{f_{i}^{(1)}, \dots, f_{i}^{(K)}\} \leftarrow \pi_{\theta}(m_k^{(i-1)}\oplus q_i \oplus a_{i}^{(k)} \oplus r_{i}^{(k)}) \text{ for } k=1 \dots K$ \text{\,\,\,\,\,\,\,\,// Generate feedbacks.}
        \STATE $m_k^{(i)} \leftarrow m_k^{(i-1)} \oplus (q_i, a_i^{(k)}, r_{i}^{(k)}, f_i^{(k)})$ \text{ for } $k=1 \dots K$
        \text{\,\,\,\,\,\,\,\,// Update in-context memory.}
    \ENDFOR
    
    \STATE Construct final prompts: $\mathcal{P} \leftarrow \{ m_k^{(N)} \cup \{x_{\text{test}}\} \}_{k=1}^K$
    \STATE $\{ \hat{a}_*^{(1)}, \dots, \hat{a}_*^{(K)} \} \leftarrow \text{Parallel}(\pi_{\theta}(p) \text{ for } p \in \mathcal{P})$
    
    \STATE $y^* \leftarrow \text{Vote}(\{ \text{Extract}(\hat{a}_*^{(k)}) \}_{k=1}^K)$
    \RETURN $y^*$
\end{algorithmic}
\end{algorithm*}

Our framework, illustrated in Algorithm~\ref{alg:TR-ICRL}, comprises three distinct phases designed to iteratively refine model reasoning through contextual alignment:
(1) Context Retrieval: Given a target question, we retrieve the most semantically relevant questions from the unlabeled evaluation set to serve as in-context demonstrations for TR-ICRL (Section~\ref{sec:retrieval}).
(2) Test-Time Iterative Rethinking: For each retrieved query, the model generates an initial reasoning trajectory. Initial trajectories are evaluated against pseudo-labels derived via majority voting.
This initiates a rethinking stage where the LLM receives reward messages and generates feedback, enabling it to recursively update its in-context memory. (Section \ref{sec:thinking}).
(3) Self-Consistent Aggregation: The reasoning outputs from all contexts are aggregated, and the final answer is determined by majority voting (Section~\ref{sec:voting}).

\subsection{Context Retrieval}
\label{sec:retrieval}
Unlike traditional Reinforcement Learning, TR-ICRL bypasses weight updates via backpropagation in favor of gradient-free policy optimization. This is achieved by iteratively refining the model’s local context during inference. Under this paradigm, unlabeled instances retrieved from the evaluation set serve as the primary substrate for unlabeled ICRL. These instances provide the necessary contextual foundation for the model to autonomously refine its policy behavior without the need for ground-truths.

To ensure high task-relevance, we employ an embedding model to vectorize the input queries. We then compute the cosine similarity between the target question and the unlabeled set to retrieve the most semantically similar instances. These retrieved cases serve as the initial prompts for the Test-Time Iterative Rethinking process.

\subsection{Test-Time Iterative Rethinking}
\label{sec:thinking}

Inspired by self-consistency with CoT~\cite{wang2022self}, which shows that correct answers tend to form dense and consistent clusters among multiple model outputs, we generate multiple predictions for the retrieved questions and then apply majority voting to get the pseudo-labels.
These labels serve as a ground-truth proxy during the rethinking stage and generates reward messages based on its correctness.
By incorporating these rewards back into the context, the LLM learns from its prior outputs.
This allows the model to systematically identify reasoning flaws and develop a more comprehensive understanding of the problem at inference time.
We provide a detailed description of the overall process below.

\paragraph{Rollout}
Given the retrieved questions from the context retrieval stage, LLM generates the multiple predictions, in the format of zero-shot CoT reasoning~\cite{kojima2022large}.

For each retrieved question $\hat{x_i}$, the LLM input $\alpha_i$ is formatted as:
\begin{equation}
\alpha_i = \text{Q: } \hat{x_i}. \text{ A: } \texttt{[Z]},
\end{equation}
where $i$ represents the $i$-th step, and \texttt{[Z]} represents zero shot trigger~\cite{guo2025deepseek}.
More details about the triggers used for different benchmarks are described in Appendix~\ref{sec:ict}.
Subsequently, based on the input $\alpha_i$, LLM is instructed to generate the prediction for the retrieved question, obtaining the ${y}_i$.

Furthermore, recognizing that some LLMs may not fully comply with the instructions when answering a query (e.g., by refusing to answer), TR-ICRL incorporates an additional filtering step to exclude abnormal responses that deviate from the given instructions.

\paragraph{Rethinking}
The rethinking stage in TR-ICRL consists of two distinct processes: \textbf{reward} and \textbf{feedback}.
During the reward process, we use majority voting to get a pseudo-label \({y}^*\), \textit{i.e.},
\begin{equation}
    {y}^* = \arg\max_{c \in \mathcal{C}} \sum_{i=1}^{i} \mathbf{1}\{{y}_i = c\}.
\end{equation}

Then the prediction \(y_i\) is evaluated against the corresponding the pseudo-label \({y}^*\), which can be given as:
\begin{equation}
R_i =
    \begin{cases}
    R_{\text{correct}}, & \text{if } {y}^* = {y}_i,\\[2mm]
    R_{\text{wrong}}, & \text{otherwise.}
    \end{cases}
\end{equation}
Specifically, we assign a binary reward based on correctness (e.g., 'Well done! Your answer is correct.')."

While existing studies~\cite{dai2022can} establishes a theoretical duality between in-context learning and fine-tuning, a critical functional discrepancy remains: fine-tuning explicitly computes gradients based on the divergence between predictions and ground-truth labels.

To bridge this gap, we introduce a feedback process wherein the LLM generates a corrective response $f_i$, conditioned on a reward message.
This process is designed to emulate the weight-adjustment logic of a gradient update; specifically, it lets the model rethink its prior output through the reward message. 
For correct predictions, this feedback reinforces the established logical trajectory. Conversely, for incorrect instances, it facilitates error diagnosis and rectifies flawed intermediate reasoning steps.

\subsection{Self-Consistent Aggregation}
\label{sec:voting}
The original question is appended to the context messages, and the combined messages are presented to the model.
The model generates the final responses from the enriched context messages, enabling reasoning guided by the prediction to similar questions and the associated reward messages. Each generated response can be given as:
\begin{equation}
    {A_i} = \text{LLM}(D_i, x).
\end{equation}
In this context, $D_i$ refers to the context messages in $i$-th time sample.
Finally, the model generates $N$ corresponding responses to the original question.
Each generated response is considered a candidate answer, $ A = \{A_1, A_2, \dots, A_N\} $.
The final answer ${A^*}$ is determined by majority voting, selecting the candidate that appears most frequently.

\section{Experiments}
\subsection{Experimental Setup}

\begin{figure*}[t!]
\centering
\includegraphics[width=0.90\textwidth]{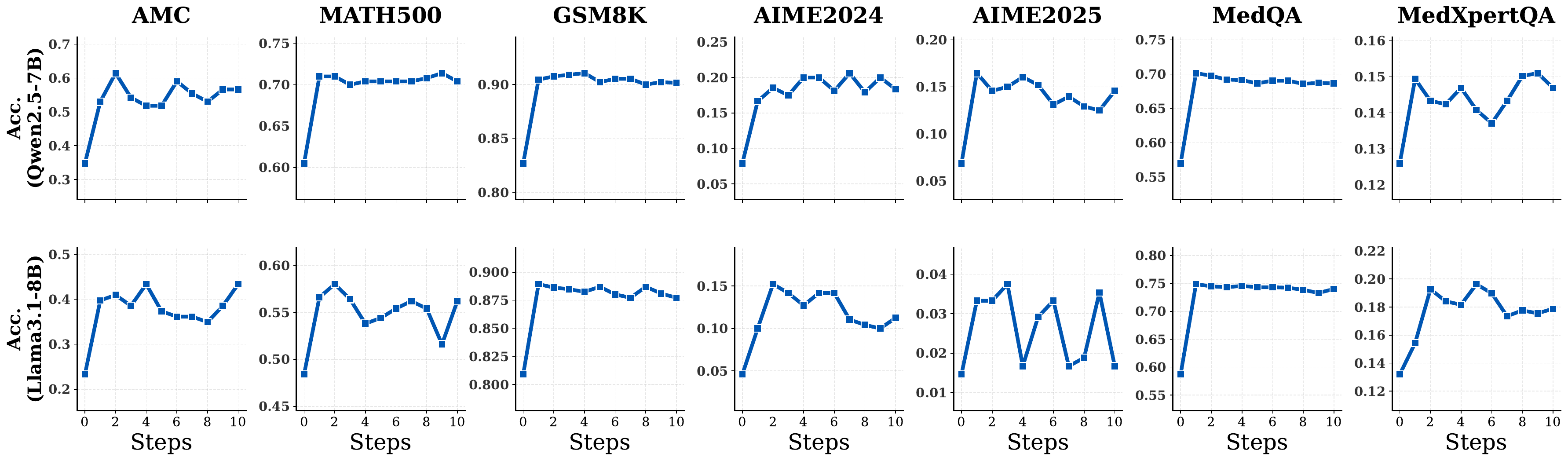}
\caption{We evaluate \textbf{TR-ICRL} across a range of 1 to 10 ICRL steps within both reasoning and knowledge-intensive tasks. Performance at step 0 serves as the experimental baseline.}
\label{fig:main_exp}
\end{figure*}

\paragraph{Models} 
To evaluate the generality of TR-ICRL across different backbones, we conduct experiments using Qwen2.5-7B-Instruct \cite{yang2024qwen2} and Llama3.1-8B-Instruct \cite{grattafiori2024llama} as instruct-tuned models.
For large reasoning models, we employ Qwen3-8B \cite{qwen3} and DeepSeek-R1-0528-Qwen3-8B \cite{guo2025deepseek}.

\paragraph{Benchmarks}
To evaluate the applicability of TR-ICRL across reasoning and knowledge-intensive tasks of varying difficulty, we assess its performance on three widely used reasoning benchmarks: MATH500~\cite{hendrycks2021measuring}, AMC~\cite{li2024numinamath}, and GSM8K~\cite{cobbe2021training}, as well as on two more challenging reasoning benchmarks, AIME2024\footnote{\url{https://huggingface.co/datasets/HuggingFaceH4/aime_2024}}
and AIME2025\footnote{\url{https://huggingface.co/datasets/opencompass/AIME2025}}.
To assess performance on knowledge-intensive tasks, we evaluate TR-ICRL using MedQA~\cite{jin2021disease}, a standard medical benchmark.
Furthermore, we extend our evaluation to MedXpertQA\footnote{\url{https://huggingface.co/datasets/TsinghuaC3I/MedXpertQA/tree/main/Text}}, a challenging medical knowledge benchmark.

\paragraph{Implementation Details}

We employ vLLM \cite{kwon2023efficient} for online inference, deploying the model on $2$*NVIDIA A100 ($80$GB) GPUs.
For context retrieval, we utilize Qwen3-8B-Embedding \cite{zhang2025qwen3} to generate vector representations.
During inference, we sample responses using a temperature of $0.6$ and $top\_p=0.8$, with the maximum number of generated tokens to $8192$.
For each retrieval question, we sample $8$ times to get pseudo-label via majority voting.
In the majority voting, we select the answer that appears most frequently as the final prediction. If there are multiple options with the same frequency, we randomly select one as the final answer.
We use accuracy as the evaluation metric.

\subsection{Main Results}

\paragraph{TR-ICRL performs well on most tasks and models}
TR-ICRL achieves robust and significant improvements across various benchmarks.
The experimental results, detailed from \textbf{step 1} through \textbf{step 10} in Figure \ref{fig:main_exp}, demonstrate the scalability and consistency of our approach.
on common reasoning benchmarks, Qwen2.5-7B with TR-ICRL improves performance by \textbf{16.72\%} on MATH500 and \textbf{58.91\%} \textbf{(from 34.80 to 55.30)} on AMC, demonstrating consistent gains on standard mathematical reasoning tasks.
More notably, on the more challenging reasoning benchmarks, TR-ICRL leads to dramatic relative improvements ranging from \textbf{137.59\%} \textbf{(from 7.90 to 18.77)} on AIME2024 and \textbf{109.84\%} \textbf{(from 6.88 to 14.43)} on AIME2025 for Qwen2.5-7B.
These results indicate that TR-ICRL is particularly effective at enhancing complex, multi-step reasoning capabilities where base models struggle most.

Beyond reasoning tasks, TR-ICRL also exhibits strong generalization to knowledge-intensive tasks.
On the MedQA, LLaMA3.1-8B equipped with TR-ICRL outperform their respective backbones by \textbf{26.43\%}.
On more challenging medical benchmark MedXpertQA, Llama3.1-8B with TR-ICRL surpasses the backbone by \textbf{36.68\%} \textbf{(from 13.20 to 18.04)}.
These results underscore the broad applicability and robustness of TR-ICRL, demonstrating its effectiveness across both reasoning and knowledge-intensive tasks.

However, we observe a performance degradation during the latter stages of TR-ICRL.
In the iterative rethinking process, the context serves as a dynamic buffer that accumulates task instructions, multiple rollouts, reward signals, and feedback. 
As this sequence length expands, the model becomes increasingly susceptible to informational interference.
Specifically, the model struggles to maintain contextual focus when reconciling contradictory or noisy data pairs within a long-form sequence.
This distraction from accumulating history can dilute the signal of the most recent rewards, leading to a breakdown in the model's ability to internalize the optimal reasoning path.

\paragraph{TR-ICRL Performs well on LRMs}
\begin{figure}[t]
  \centering
  \includegraphics[width=\columnwidth]{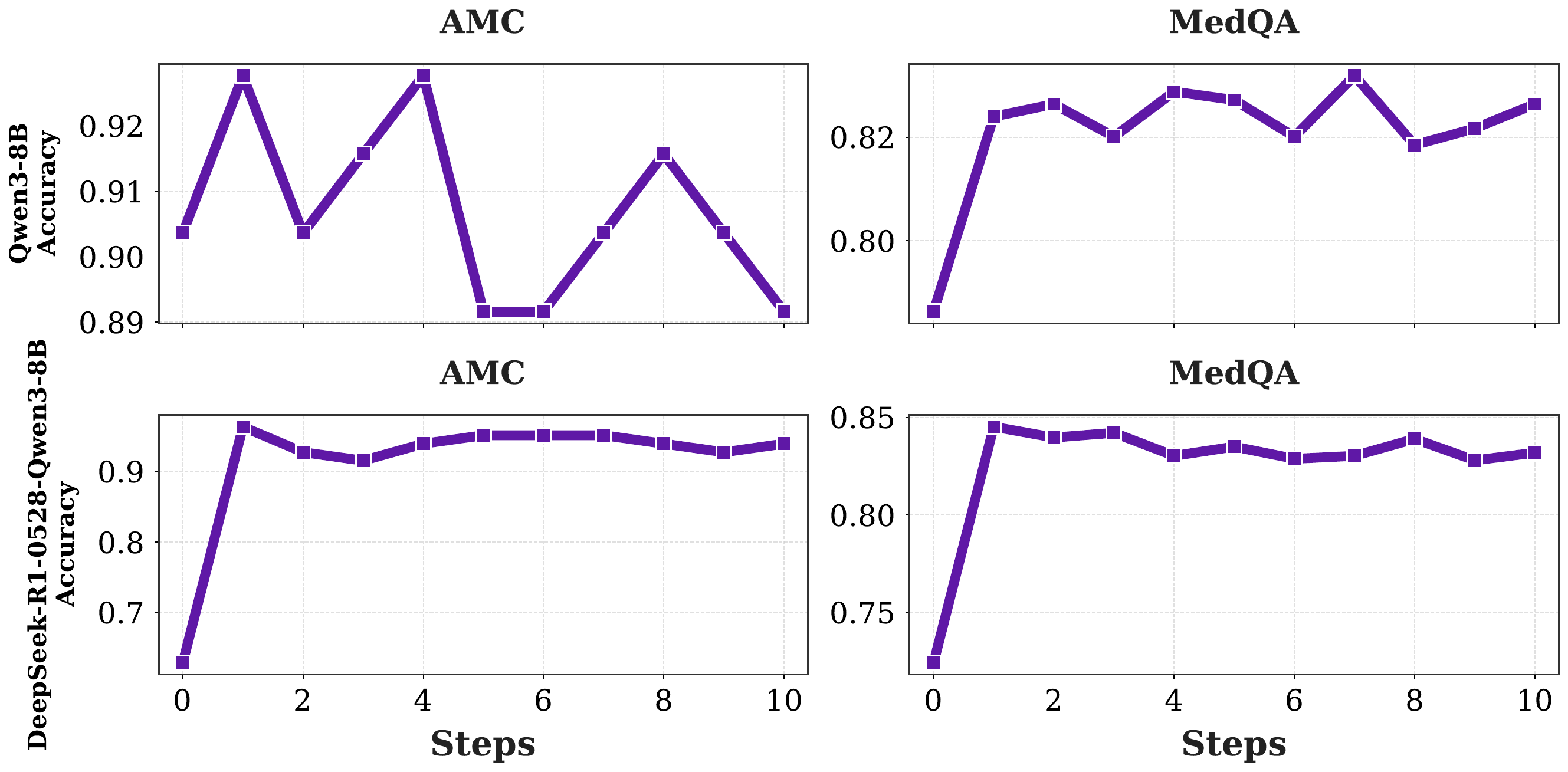}
   \caption{The evolution of LRM performance.}
  \label{fig:lrm_cure}
\end{figure}

LRMs are increasingly becoming central to contemporary research and applications.
We thus conducted experiments to evaluate the effectiveness of TR-ICRL on LRMs.
Our results demonstrate that LRMs achieve significant improvements by integrating their inherent reasoning capabilities with TR-ICRL.
As illustrated in Figure~\ref{fig:lrm_cure}, DeepSeek-R1-0528-Qwen3-8B exhibits a remarkable \textbf{49.79\%} \textbf{(from 62.82 to 94.09)} improvement on the AMC benchmark. Furthermore, this approach extends effectively to the medical domain; on the MedQA benchmark, the same model yields a significant 15.28\% increase in accuracy, underscoring the robustness of the proposed method across diverse reasoning tasks.
On Qwen3-8B, contextual interference is markedly more severe; specifically, performance at steps 6 and 10 on the AMC falls below the baseline.

\subsection{Ablation study}
To evaluate the specific impact of the retrieved question distribution, we conduct ablation studies using three alternative selection strategies:
(1) Random: We select questions by randomly sampling from the unlabeled evaluation set, rather than relying on contextual similarity.
(2) Min-Similarity: We intentionally retrieve questions with the lowest similarity scores to test the boundaries of relevance.
(3) Cross-Domain: We select questions from unrelated fields to assess the robustness of TR-ICRL when retrieved examples originate from a different distribution.
For instance, we utilize MedQA (medical domain) as the retrieval source for mathematical problems and MATH500 (math domain) as the source for medical queries.
The results are summarized in Figure~\ref{fig:ablation}, with the detailed descriptions are provided in Appendix~\ref{sec:more_ablation}.
\begin{figure*}[t!]
\centering
\includegraphics[width=0.80\textwidth]{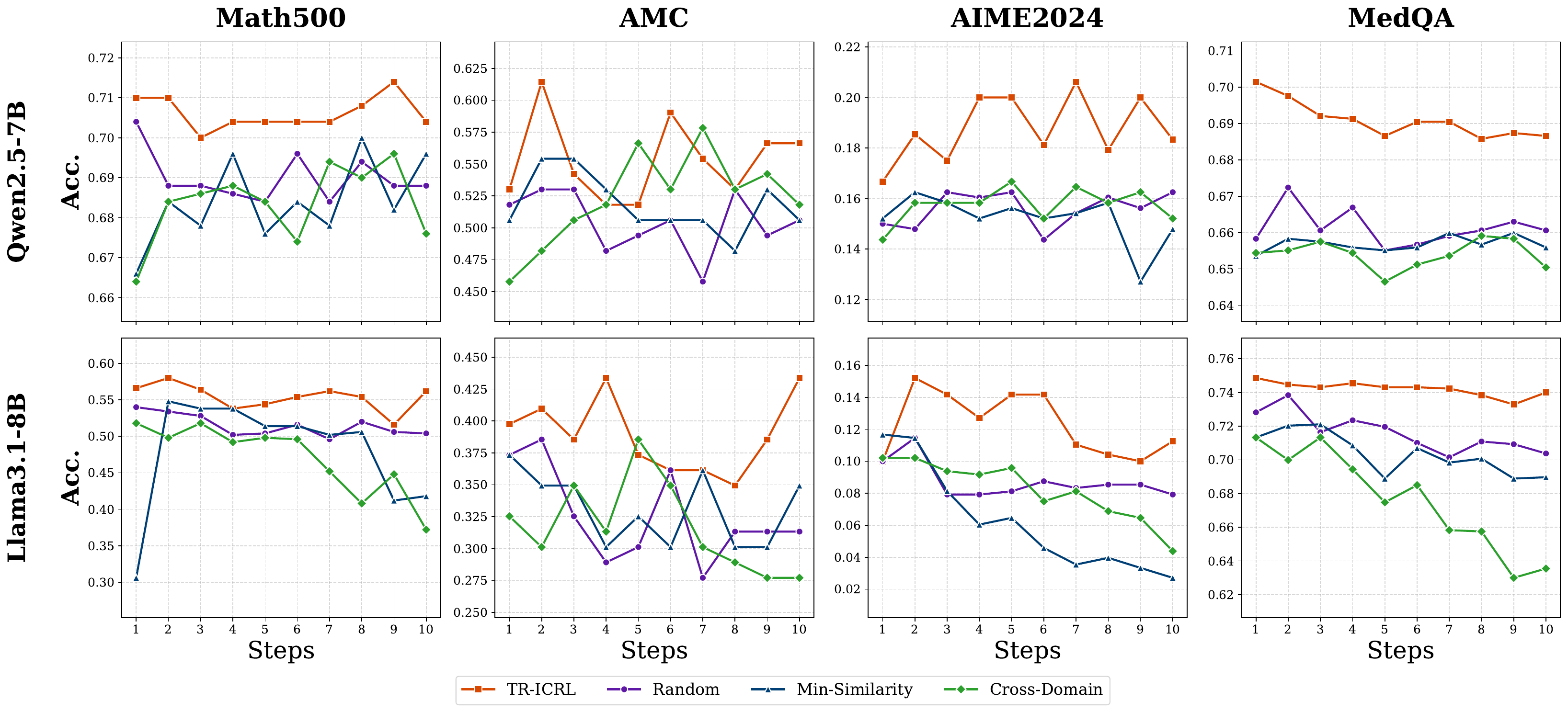}
\caption{Ablation study of retrieved question distribution in TR-ICRL.}
\label{fig:ablation}
\end{figure*}
Min-similarity and random configurations consistently underperformed.
This performance drop underscores the critical role of context relevance in ICRL.
Cross-domain configuration highlights a clear limitation in TR-ICRL regarding out-of-distribution generalization; specifically, the guidance provided by cross-domain cases is demonstrably less effective than that derived from in-domain.

\section{Analysis}
\begin{figure*}[t]
  \centering
  \includegraphics[width=0.80\textwidth]{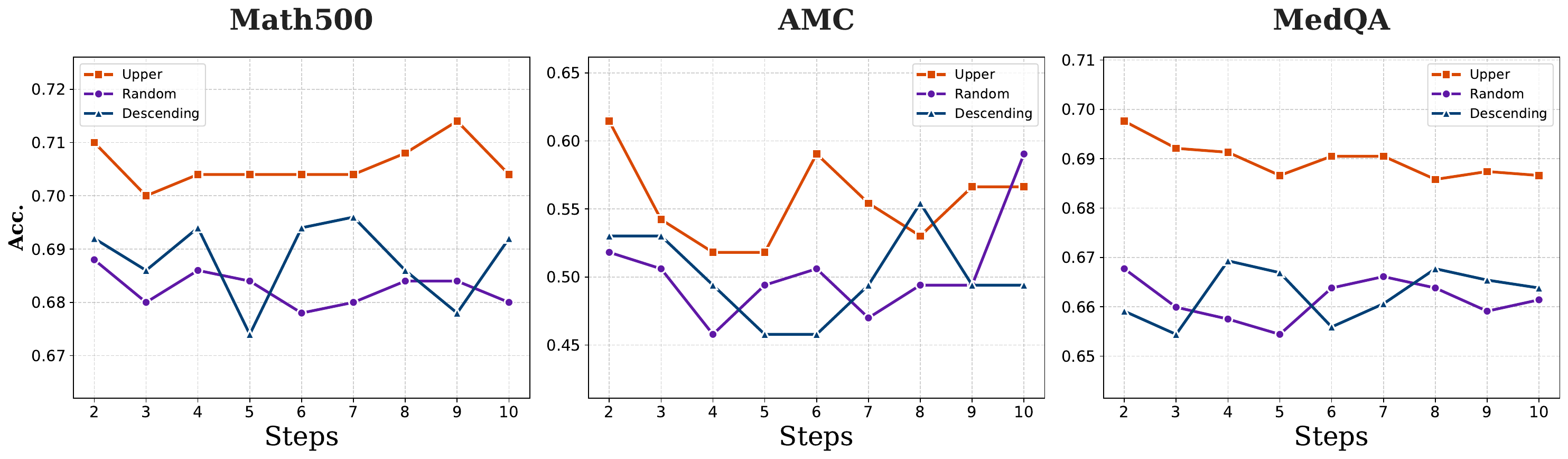}
   \caption{Performance impact of contextual sequence ordering across diverse benchmarks.}
  \label{fig:sequence}
\end{figure*}

\subsection{Contextual coherence in TR-ICRL}
To evaluate the impact of contextual relevance on model performance, we conducted a series of experiments varying the presentation order of retrieved content.
We compared three distinct configurations—increasing similarity, decreasing similarity, and randomized ordering, as illustrated in Figure \ref{fig:sequence}.

Across all three datasets, the \textbf{increasing} configuration consistently outperformed both the decreasing and random strategies.
On Math500, the increasing strategy maintained a significant margin, achieving a peak accuracy of $71.4\%$ at step 9.
This suggests that the model’s reasoning efficacy is enhanced when highly relevant information is presented in closer proximity to the query prompt.
Conversely, the decreasing strategy exhibited lower accuracy, most notably on the AMC dataset where performance reached a nadir of approximately $0.46$ at step 5.
The performance fluctuations observed in the decreasing configuration across datasets indicate that distancing highly relevant context from the final instruction introduces noise and degrades reasoning quality.
These findings demonstrate that for TR-ICRL, prioritizing the proximity of relevant information to the final reasoning step is a critical design choice for maximizing performance.

\subsection{Why TR-ICRL work in challenging benchmarks?}
We observed a remarkable phenomenon: on challenging benchmarks such as AIME 2024, the base model initially achieves a score of only \textbf{7.9}.
However, without any additional fine-tuning, it reaches significantly higher performance through TR-ICRL.
We attribute this leap to a reward mechanism based on majority voting, a phenomenon we define as the '\textbf{lucky reward}'.

\begin{figure}[t]
  \centering
  \includegraphics[width=\columnwidth]{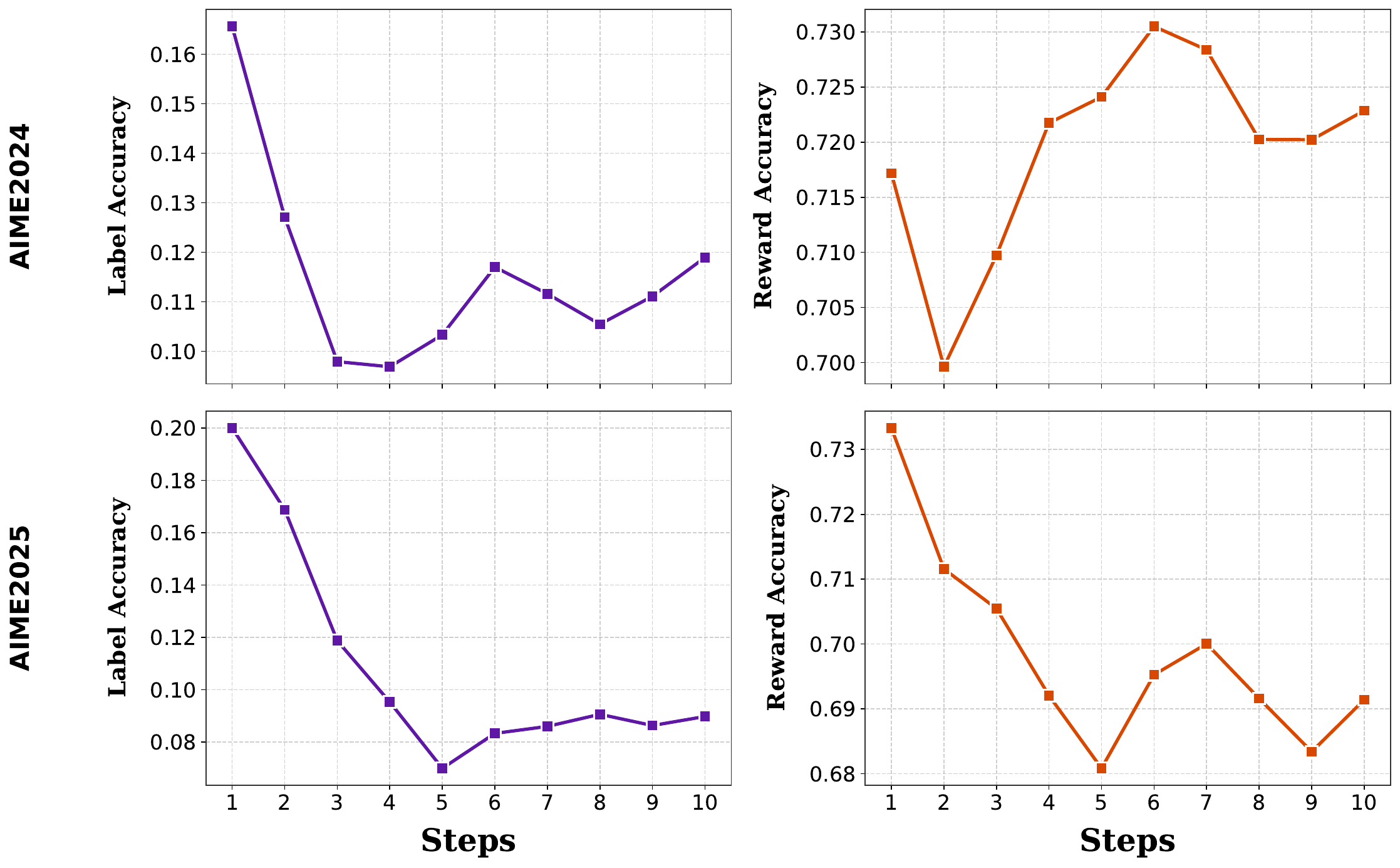}
   \caption{Comparison of label accuracy and reward accuracy on AIME2024 and AIME2025. Even with low label accuracy, reward accuracy remains high due to “lucky reward”.}
   \vspace{-1.2em}
  \label{fig:why_work}
\end{figure}

For an incorrectly predicted answer, even if the estimated label does not match the ground truth label, as long as it differs from the predicted answer, we will still output a negative reward.
Namely, it is sufficient that the estimated label differs from the predicted answer for us to assign the correct negative reward.
Reward messages are denser than labels, allowing for more opportunities to get useful reward messages even when the estimated label is inaccurate.

Therefore, we find that the majority voting rewards in TR-ICRL remain remarkably accurate even as model capability decreases.
As shown in Figure~\ref{fig:why_work}, on AIME2024, for instance, while majority voting achieves a raw accuracy of only \textbf{16.56\%}, the resulting reward accuracy reaches \textbf{71.71\%}.
A similar trend appears in AIME2025, where a low accuracy of \textbf{20.00\%} still yields a reward accuracy of \textbf{73.33\%}.

\textbf{'Lucky Reward'} ensures that most outputs receive correct reinforcement despite inaccurate label estimation.
While poorer model performance leads to more frequent mistakes, it also raises the likelihood that an estimated label will accurately flag those mistakes as incorrect. This phenomenon ensures a consistently high reward accuracy, creating a reliable signal that supports effective learning during the test time.

\subsection{Do reward messages help?}

\begin{figure}[t]
  \centering
  \includegraphics[width=\columnwidth]{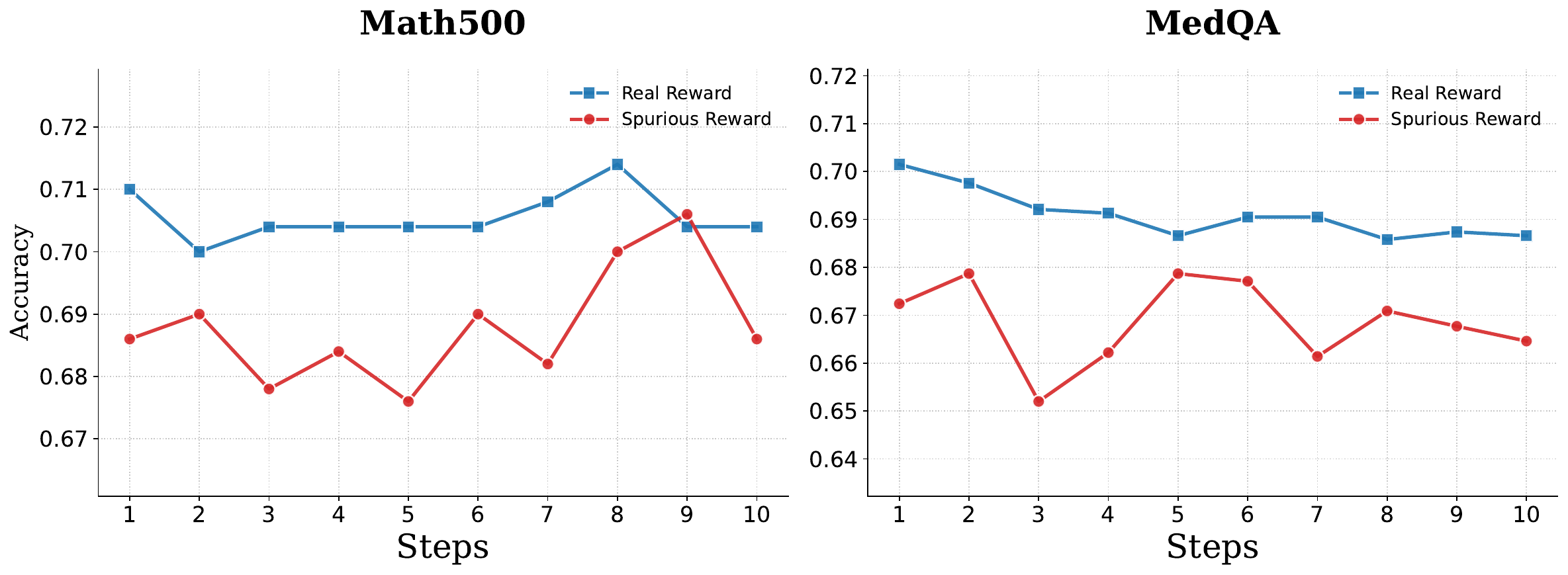}
   \caption{Results of spurious rewards on Qwen2.5-7B.}
   \vspace{-0.5em}
  \label{fig:fake_reward}
\end{figure}

Inspired by ~\cite{shao2025spurious}, we wanted to explore the effectiveness of reward messages in ICRL. We also conducted a spurious reward experiment on Qwen2.5-7B.

As shown in Figure~\ref{fig:fake_reward}, we observe a consistent performance gap between TR-ICRL and the spurious rewards setting across all configurations.
Under the real reward setting, the model maintains a relatively high accuracy, fluctuating between 0.70 and 0.715.
Conversely, the spurious reward consistently results in lower accuracy, generally staying below 0.69.
Notably, at step 9, the spurious reward performance briefly converges with the real reward, before sharply declining again at step 10.
The performance gap is even more pronounced in the knowledge-intensive tasks.

These results highlight the critical role of reward message correctness.
When rewards aligned with the majority consensus, the model is encouraged to consolidate reliable reasoning trajectories.
In contrast, spurious reward introduces systematic inconsistencies between correct answers and feedback signals, thereby distorting the learning signal and limiting performance improvements.

Interestingly, the spurious reward setting still outperforms the vanilla baseline. We attribute this phenomenon to several factors.
Integrating the feedback into the context messages facilitates a more effective dual descent gradient.
This approach establishes a more precise dual formulation that bridges In-Context Learning and fine-tuning.
For relatively simple questions, the model often derives the correct answer based on prior knowledge or common-sense reasoning.
Even when prompted to revise its reasoning, the model tends to converge to the same conclusion, meaning that spurious rewards have minimal impact on the final prediction.

In contrast, for sufficiently difficult questions where the model consistently fails to produce correct answers, spurious rewards actually tell them that they answered incorrectly and help the model reconsider its reasoning trajectory.

\subsection{Comparisons with other methods}
To evaluate its efficiency, we compare the performance of TR-ICRL (Step 1) against several established baselines, such as Best-of-N~\cite{brown2024large}, Self-Refine~\cite{madaan2023self} and Reflexion~\cite{shinn2023reflexion}, under a \textbf{fixed token} budget.
These results are summarized in Table~\ref{tab:tts}.

TR-ICRL consistently achieves superior performance across the majority of evaluated benchmarks, notably outperforming the BoN baseline. 
These results suggest that TR-ICRL provides a more disciplined and robust framework for complex reasoning trajectories. 
Unlike BoN, which depends on diversified sampling, TR-ICRL implements a structured feedback loop. 
Furthermore, Self-Refine is often limited by the cognitive bottleneck of internal verbal critiques, particularly in high-complexity math tasks.
TR-ICRL leverages informative contextual anchors, such as historical predictions and feedbacks, by internalizing these corrective cues through ICRL, the model generates more reliable and logically coherent inference paths.
Even in the AMC dataset, where Reflexion achieves a peak of 55.42\%, TR-ICRL remains highly competitive while demonstrating much higher robustness on the more challenging math tasks.
Ultimately, TR-ICRL’s ability to yield substantial accuracy gains without sacrificing computational efficiency underscores its practical superiority and robust potential for complex downstream reasoning tasks.

\begin{table}[t]
\centering
\footnotesize
\setlength{\tabcolsep}{3pt}
\renewcommand{\arraystretch}{1.1}
\caption{Comparison of TR-ICRL with BoN, Self-Refine and Reflexion.}
\label{tab:tts}
\scalebox{0.9}{
\begin{tabular}{lcc}
    \toprule
    \textbf{Setting} & \textbf{Acc.(MATH)} & \textbf{Acc.(AMC)} \\
    \midrule
    \textbf{TR-ICRL (Step 1)} & \textbf{71.00} & 53.01 \\
    BoN (N=16) & 69.60 & 43.37 \\
    Self-Refine (Iteration=8) & 61.20 & 42.17 \\
    Reflexion & 66.60 & \textbf{55.42} \\
    \bottomrule
\end{tabular}
}
\end{table}

\section{Conclusion}
In this paper, we propose \textbf{Test-Time Rethinking for In-Context Reinforcement Learning (TR-ICRL)}, a novel framework for ICRL on test time without access to ground-truth labels.
A central innovation of TR-ICRL is its rethinking stage, which is driven by two integrated processes: reward estimation and feedback generation.
In the reward process, we leverage majority voting to derive a reliable reward.
Then, the model refines its reasoning by incorporating reward messages, achieving autonomous optimization without requiring external intervention.
Our experiments demonstrate the strong potential of TR-ICRL, achieving consistent improvements across a variety of models and tasks.
As a result, TR-ICRL presents itself as a promising method for ICRL.

\newpage
\section*{Limitations}
Despite its effectiveness, TR-ICRL’s reliance on a simple majority vote in reward estimation may be overly reductive.
Moreover, this reward estimation generates a purely binary reward signal, it lacks the nuance required for complex reasoning tasks.
To enhance decision making robustness, it is essential to differentiate between various inference paths. Incorporating uncertainty metrics, such as Perplexity (PPL) or Entropy would allow the framework to quantify the confidence of each trajectory.
Furthermore, replacing binary signals with a rubric-based reward system would enable more granular scoring.

% Bibliography entries for the entire Anthology, followed by custom entries
%\bibliography{anthology,custom}
% Custom bibliography entries only
\bibliography{custom}

\newpage
\clearpage

\appendix

\section{TR-ICRL Implementation Details}
\label{sec:ict}

\subsection{Question prompt template}

In MedQA, we use zero-shot CoT template adapted from Deepseek-R1:
\begin{tcolorbox}[
    enhanced,
    colback=cyan!10!white,
    colframe=black,
    coltext=black,
    boxrule=1pt,
    arc=3mm,
    breakable
]
\texttt{Q: \{question\}\textbackslash nA: Please reason step by step, and put your final answer (selected from options A to D) within \textbackslash boxed\{\}.}
\end{tcolorbox}

In MedXpertQA, because the range of answer is different, this is another template:
\begin{tcolorbox}[
    enhanced,
    colback=cyan!10!white,
    colframe=black,
    coltext=black,
    boxrule=1pt,
    arc=3mm,
    breakable
]
\texttt{Q: \{question\}\textbackslash nA: Please provide a step-by-step explanation, followed by your final answer (selected from options A to J) within \textbackslash boxed\{\}.}
\end{tcolorbox}

In reasoning benchmark, we will use the following template to guide the responses:
\begin{tcolorbox}[
    enhanced,
    colback=cyan!10!white,
    colframe=black,
    coltext=black,
    boxrule=1pt,
    arc=3mm,
    breakable
]
\texttt{Q: \{question\}\textbackslash nA: Please reason step by step, and put your final answer within \textbackslash boxed\{\}.}
\end{tcolorbox}

\subsection{Reward template}
After the model generates a prediction for a retrieval question, it is verified against the corresponding the pseudo-label.
If the prediction is correct, a positively reward label is appended to the context, affirming the validity of the reasoning process.
In cases of incorrect predictions, instead of explicitly pointing out the error, a supportive and constructive reward label is introduced. This approach encourages further reflection and exploration without directly identifying the mistake.\\
When the prediction is correct:
\begin{tcolorbox}[
    enhanced,
    colback=cyan!10!white,
    colframe=black,
    coltext=black,
    boxrule=1pt,
    arc=3mm,
    breakable
]
\texttt{User: Well done! Your answer is correct.}
\end{tcolorbox}

When the prediction is wrong:
\begin{tcolorbox}[
    enhanced,
    colback=cyan!10!white,
    colframe=black,
    coltext=black,
    boxrule=1pt,
    arc=3mm,
    breakable
]
\texttt{User: Unfortunately, your answer is wrong! Review your previous answer. Find the reason for the mistake.}
\end{tcolorbox}

\section{A Detail ablation study}
\label{sec:more_ablation}
In ablation analysis, we selected MATH500, AMC and AIME2024 as the reasoning benchmarks, and MedQA as the knowledge-intensive benchmark.

The ablation study reveals a significant performance gap when the quality of the retrieved context is compromised.
Both the random and min-similarity configurations consistently underperform relative to the standard TR-ICRL framework across all benchmarks.
For a model to effectively utilize in-context reinforcement, the retrieved questions must be mathematically or logically relevant to the target problem to provide meaningful guidance during the iterative thinking process.
While random sampling bypasses similarity entirely, min-similarity intentionally selects the least relevant cases; both strategies fail to match the performance gains achieved when contexts are selected based on high similarity scores.

The cross-domain strategy consistently yields the lowest performance across nearly all benchmarks, often trailing significantly behind the standard TR-ICRL.
This performance gap is particularly pronounced in the AIME2024 and MedQA tasks, where accuracy drops sharply when the model is provided with out-of-distribution contexts.
These results highlight a clear limitation in OOD generalization; the reasoning patterns inherent in one domain (e.g., medical knowledge) do not effectively translate to provide helpful guidance for another (e.g., competitive mathematics).
The data demonstrates that guidance derived from cross-domain cases is demonstrably less effective than in-domain examples, underscoring that the benefits of TR-ICRL are highly dependent on domain-aligned context.

\begin{table*}[t]
\centering
\small
\setlength{\tabcolsep}{8pt}
\begin{tabular}{lcccccccc}
\toprule
{\textbf{Model}} & \textbf{Params} & \multicolumn{2}{c}{\textbf{Avg. Tokens ($k$)}} & \multicolumn{2}{c}{\textbf{Relative Cost (units)}} & \multicolumn{2}{c}{\textbf{Accuracy (\%)}} \\
\cmidrule(lr){3-4} \cmidrule(lr){5-6} \cmidrule(lr){7-8}
& \textbf{(B)} & MATH & AMC & MATH & AMC & MATH & AMC \\
\midrule
Qwen2.5-72B & 72 & 0.62 & 0.92 & 89.28 & \textbf{132.48} & 62.10 & 41.11 \\
TR-ICRL (step 1) & 7 & \textbf{0.48} & \textbf{0.81} & \textbf{80.64} & 136.08 & \textbf{71.00} & \textbf{53.01} \\
\bottomrule
\end{tabular}
\caption{Resource vs. Performance Comparison on MATH and AMC Benchmarks. We report the total computational overhead using RCC. For the Qwen2.5-72B baseline, cost is calculated based on \textbf{Best-of-2} majority voting. In contrast, TR-ICRL (Step 1) achieves superior accuracy using a single iterative step within Qwen2.5-7B.}
\label{tab:resource_performance}
\end{table*}

\section{TR-ICRL Is comparable to or Outperforms Large Parameter LLMs}

To ensure a fair and rigorous comparison between the performance of smaller models (7B) empowered by \textbf{TR-ICRL} and larger baseline models (72B), we conducted a controlled-variable analysis focusing on total computational overhead (FLOPs)~\cite{chen2023run}. 

For a single forward pass, the total FLOPs can be approximated as $C \approx 2 \cdot P \cdot N_{\text{tokens}}$, where $P$ is the parameter count and $N_{\text{tokens}}$ is the sequence length. We define the \textbf{Relative Compute Cost (RCC)} as:
\begin{equation}
RCC = P \times \text{Avg. Tokens per Response}
\end{equation}

By fixing the computational budget, we demonstrate that TR-ICRL enhanced small models not only outperform larger models but do so with high resource efficiency.
As illustrated in Table~\ref{tab:resource_performance}, empirical results reveal that TR-ICRL yields substantial absolute accuracy gains of 14.3\% on MATH and 28.9\% on AMC over the much larger 72B baseline.
These findings validate that our iterative framework effectively bridges the performance gap between model scales, enabling a 7B-parameter model to surpass a 72B-parameter model through ICRL.

\section{Additional Experiments Details}

\subsection{Baseline Models}
For all baseline models, we use zero-shot to inference.
For Large reasoning Models, we follow the corresponding recommended prompting guidelines to remove the system prompt. \\
The zero-shot template is :
\begin{tcolorbox}[
    enhanced,
    colback=cyan!10!white,
    colframe=black,
    coltext=black,
    boxrule=1pt,
    arc=3mm,
    breakable
]
\texttt{Q: \{question\}\textbackslash nA: Put your final answer within\textbackslash boxed\{\}.}
\end{tcolorbox}

\subsection{Data Statistics}
The detailed benchmark statistics are shown in Table \ref{tab:benchmark}.

\begin{table}[!tbp]
\centering
\resizebox{0.95\columnwidth}{!}{
\begin{tabular}{lccc}
    \toprule
    \textbf{Dataset} & \textbf{Train Num} & \textbf{Test Num} & \textbf{Options Num} \\
    \midrule
    MATH500 & 0 & 500 &N/A \\
    \midrule
    AMC & 0 & 83 &N/A \\
    \midrule
    GSM8K & 7473 & 1319 &N/A \\
    \midrule
    AIME2024 & 0 & 30 &N/A \\
    \midrule
    AIME2025 & 0 & 30 &N/A \\
    \midrule
    MedQA  & 10178 & 1273 & 4 \\
    \midrule
    MedXpertQA & 5 & 2450 & 10 \\
    \bottomrule
\end{tabular}
}
\caption{The Statistics of benchmarks.}
\label{tab:benchmark}
\end{table}

\subsection{Evaluation Metrics}
We employ accuracy as our evaluation metric.
To ensure statistical robustness on the AIME 2024 and AIME 2025 datasets, we report the mean accuracy across \textbf{16} independent trials.
Performance on all remaining datasets is reported based on a single experimental trial.

\subsection{Answer cleaning}
As we guide the model in generating answers, we use the \verb|\boxed{}| format to standardize the final answer output.
However, due to differences across models, we apply various regular expressions to extract the final answer accurately, as displayed in Listing~\ref{lst:extract_boxed} below.

\subsection{Majority voting}
For each reasoning step, we extract the final answer from all $K$ rollouts.
We use a majority voting strategy to select the consensus answer.
To ensure robustness in cases of parity, ties are resolved via random selection among the most frequent candidates.
This ensures that the reward signal is grounded in the most probable collective hypothesis of the model, as displayed in Listing~\ref{lst:voting} below.

\begin{figure*}[t] 
\centering
\begin{minipage}{0.85\textwidth}
\begin{lstlisting}[
    rulecolor=\color{black},
    caption={Implementation of the boxed answer extraction function.}, 
    label={lst:extract_boxed},
    abovecaptionskip=2pt,
    belowcaptionskip=7pt,
    language=Python,
    frame=single,
    basicstyle=\small\ttfamily,
    breaklines=true,
    columns=fullflexible,
    xleftmargin=0em,
    showstringspaces=false
]
def extract_boxed_answer_r1(text):
    if text is None or len(text) == 0:
        return None
    if len(text) == 1:
        return text
    match = re.search(r'\\boxed{((?:[^{}]|\{[^{}]*\})*)}', text)
    if match:
        inner_text = match.group(1)
        if len(inner_text) == 0:
            return None
        elif len(inner_text) != 1:
            text_match = re.search(r'\\text\{([A-Za-z])\}', inner_text)
            if text_match:
                return text_match.group(1)
            else:
                return inner_text
        else:
            return inner_text
    else:
        match = re.search(r'\\boxed{(.*)}', text)
        if match:
            inner_text = match.group(1)
            if inner_text.startswith('(') and inner_text.endswith(')'):
                inner_text = inner_text[1:-1]
            return inner_text
    
    answer_match = re.search(
        r'(?:Final\s+)?Answer\s*:\s*\(?([A-Z])\)?', text, re.IGNORECASE)
    if answer_match:
        return answer_match.group(1)
    return None
\end{lstlisting}
\end{minipage}
\end{figure*}

\subsection{Details in baseline Methods}
Best-of-N is implemented using OpenR~\cite{wang2024openr}. For reasoning tasks, we employ Math-Shepherd-Mistral-7B-PRM~\cite{wang2024math} as the process reward model (PRM). 

For Best-of-N, we set the temperature to 0.6, generate 16 candidate sequences with a maximum of 4096 new tokens, and select the final prediction via majority voting.

For Reflexion, we evaluate the model's output by performing multiple sampling and applying a majority voting mechanism.

\begin{figure*}[t] 
\centering
\begin{minipage}{0.85\textwidth}
\begin{lstlisting}[
    rulecolor=\color{black},
    caption={Implementation of majority voting.}, 
    label={lst:voting},
    abovecaptionskip=2pt,
    belowcaptionskip=7pt,
    language=Python,
    frame=single, 
    basicstyle=\small\ttfamily,
    breaklines=true,
    columns=fullflexible,
    xleftmargin=0em,
    showstringspaces=false
]
def vote(choices) -> str:
    if not choices:
        return None
    frequency = Counter(choices)
    max_count = max(frequency.values())
    candidates = [key for key, value in frequency.items() if value == max_count]
    result = random.choice(candidates)
    return result
\end{lstlisting}
\end{minipage}
\end{figure*}

\end{document}